\documentclass{article}

\usepackage[preprint]{neurips_2025}

\usepackage[utf8]{inputenc} 
\usepackage[T1]{fontenc}    
\usepackage{hyperref}       
\usepackage{url}            
\usepackage{booktabs}       
\usepackage{amsfonts}       
\usepackage{nicefrac}       
\usepackage{microtype}      
\usepackage{xcolor}         
\usepackage{graphicx}
\usepackage{pifont}
\usepackage{amsmath}
\usepackage{amssymb}
\usepackage{tcolorbox}
\usepackage{lipsum}
\usepackage{float}
\usepackage{cleveref}
\usepackage{comment}  

\usepackage{wrapfig}
\usepackage{colortbl}

\newcommand{\squishlist}{
\begin{list}{{{\small{$\bullet$}}}}
{\setlength{\itemsep}{3pt}      \setlength{\parsep}{1pt}
\setlength{\topsep}{1pt}       \setlength{\partopsep}{0pt}
\setlength{\leftmargin}{1em} \setlength{\labelwidth}{1em}
\setlength{\labelsep}{0.5em} } }
\newcommand{\squishend}{  \end{list}  }
\usepackage{xspace}
\newcommand{\method}{\textbf{\MakeUppercase{Reveal}}\xspace}

\graphicspath{{figures/}}

\newcommand{\sharedinternnote}{Equal contribution. Work done as Research Assistant at HKUST-GZ.}

\title{When VLMs Meet Image Classification: \\Test Sets Renovation via Missing Label Identification}

\author{
Zirui Pang$^1$\thanks{\sharedinternnote}%
\hspace{1.5em}Haosheng Tan$^2$\footnotemark[1]%
\hspace{1.5em}Yuhan Pu$^3$\footnotemark[1],\\
\textbf{Zhijie Deng}$^4$%
\hspace{1.2em}\textbf{Zhouan Shen}$^4$%
\hspace{1.2em}\textbf{Keyu Hu}$^4$%
\hspace{1.2em}\textbf{Jiaheng Wei}$^4$\thanks{Corresponding to: jiahengwei@hkust-gz.edu.cn} \\
\\
$^1$University of Illinois Urbana-Champaign \quad
$^2$University of Glasgow \quad
$^3$Boston University \\
$^4$The Hong Kong University of Science and Technology (Guangzhou)
}

\begin{document}

\maketitle

\begin{abstract}
Image classification benchmark datasets such as CIFAR, MNIST, and ImageNet, serves as critical tools for model evaluation. However, despite the cleaning efforts, these datasets still suffer from pervasive noisy labels, and often contain missing labels due to the co-existing image pattern where multiple classes appear in an image sample. This results in misleading model comparisons and unfair evaluations. Existing label cleaning methods focus primarily on noisy labels, but the issue of missing labels remains largely overlooked. Motivated by these challenges, we present a comprehensive framework named \method, integrating state-of-the-art pre-trained vision-language models (e.g., LLaVA, BLIP, Janus, Qwen) with advanced machine/human label curation methods (e.g., Docta, Cleanlab, MTurk), to systematically address both noisy labels and missing label detection in widely-used image classification test sets. \method detects potential noisy labels and omissions, aggregates predictions from various methods, and refines label accuracy through confidence-informed predictions and consensus-based filtering. Additionally, we provide a thorough analysis of state-of-the-art vision-language models and pre-trained image classifiers, highlighting their strengths and limitations within the context of dataset renovation by {\color{black}\textbf{revealing 
 10 observations}}. Our method effectively reveals missing labels from public datasets and provides soft-labeled results with likelihoods. Through human verifications, \method significantly improves the quality of \textbf{6} benchmark test sets, highly aligning to human judgments and enabling more accurate and meaningful comparisons in image classification.

\end{abstract}

\section{Introduction}
\paragraph{Noisy and Missing Labels in Image Classification Test Sets}~
Image classification is a foundational computer vision task and numerous benchmark datasets have been established to evaluate models’ visual recognition performance. However, noisy labels are occasionally present in widely used image‐classification datasets (CIFAR \citep{Krizhevsky09learningmultiple}, Caltech \citep{griffin2007caltech}, Imagenet \citep{deng2009imagenet}, MNIST \citep{lecun1998gradient}), especially in the corresponding test sets, which can compromise the reliability of model evaluations. Similar to \citep{wei2021learning,northcutt2021labelerrors}, we categorize these errors into two types \citep{northcutt2021labelerrors}: (1) \textbf{noisy label errors}, in which an incorrect label is assigned by referring to the ground-truth label \cite{natarajan2013learning,liu2016classification,wei2020optimizing,zhu2021clusterability,Wei2022ToSO}, and (2) \textbf{missing‐label errors}, where multiple object classes appear within an image but only a single label is provided as ground truth \cite{wei2021learning,northcutt2021labelerrors,wei2023aggregate}. More illustrations come as follows.

\paragraph{Caveat 1: Noisy Labels in Existing Test Benchmarks} [Termed as \emph{noisy-label} samples] 

\begin{figure}[!htb]
  \centering
  \vspace{-0.15in}
\includegraphics[width=\textwidth]{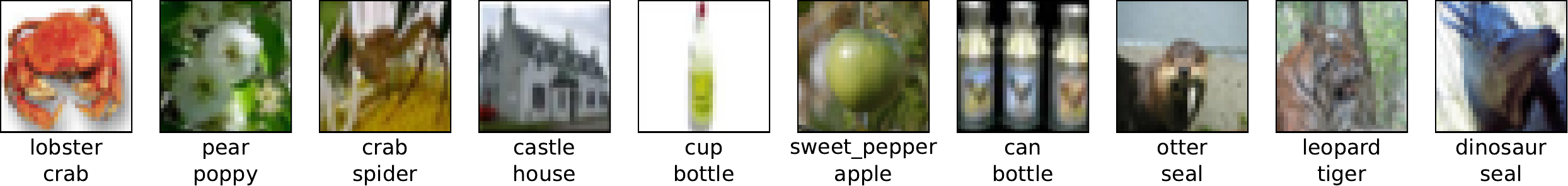}\vspace{-0.2in}
  \caption{Exemplary CIFAR-100 test set with noisy labels. The text below each picture denotes the CIFAR-100 original label (first row) and the cleaned label in CIFAR-100 by \citep{northcutt2021labelerrors} (second row).} \vspace{-0.1in}
  \label{fig: label_error_cifar100}
\end{figure}
Although benchmark test sets are intended to be precise and clean, noisy labels frequently persist, thereby undermining the integrity of evaluations. Figure \ref{fig: label_error_cifar100} illustrates several instances of such errors in the CIFAR-100 test benchmark. 
\vspace{-0.1in}
\paragraph{Caveat 2: Multi-Label Issues in Existing Test Benchmarks}[Termed as \emph{multi-label} samples] 

\begin{figure}[!htb]
  \centering
  \vspace{-0.1in}
\includegraphics[width=\textwidth]{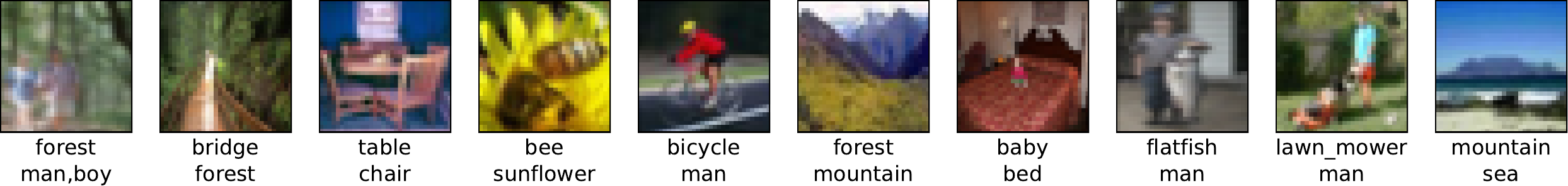}\vspace{-0.2in}
  \caption{Exemplary CIFAR-100 training images with multiple labels. The text below each picture denotes the CIFAR-100 original label (first row) and the human annotated supplementary label (second row). We did not exhaust all possible labels subjectively.}\vspace{-0.2in}
  \label{fig: multilabel error}
\end{figure}

Multi-label issues are common in image-classification benchmarks, where real-world images often contain multiple objects, but only the primary object is annotated. This leads to \textbf{co-exist} images, where valid labels are omitted, potentially biasing model evaluations. Figure \ref{fig: multilabel error} illustrates examples drawn from the CIFAR-100 test set. For instance, the first sub-figure on the left is labeled as ``forest'', yet two other classes (man, boy) which also belong to valid dataset classes, are clearly visible. Consequently, an image possibly conveys additional semantic information beyond the assigned ground-truth label, potentially biasing models evaluated on these benchmarks. 
\begin{figure}[!b]
  \centering  \includegraphics[width=1.0\textwidth]{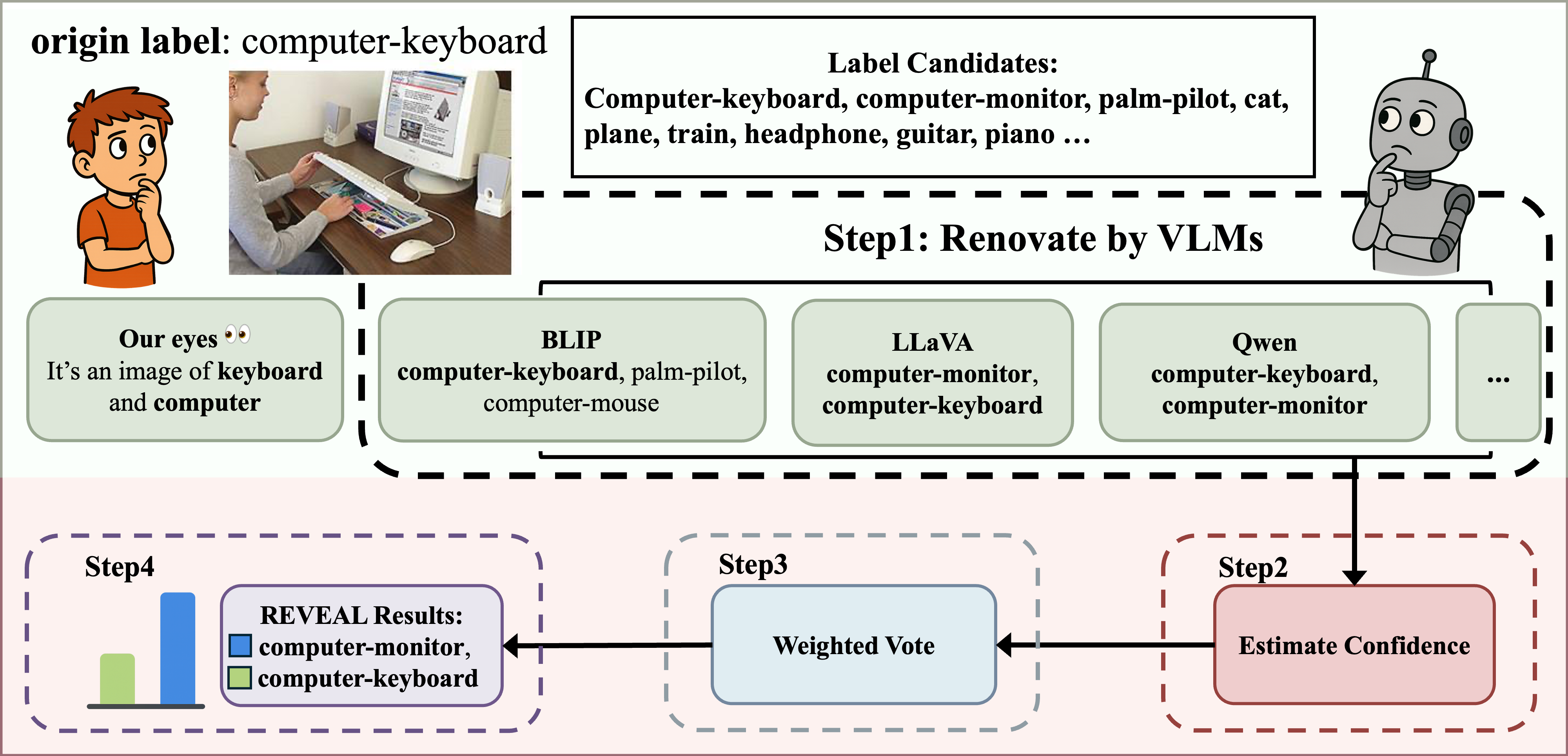}\vspace{-0.25in}
  \caption{\method renovation pipeline. Both the VLM-based and human-annotated methods first assign labels to each image independently. These preliminary labels are then aggregated using a weighted voting ensembling strategy. To refine the results, a score threshold is applied to filter the aggregated labels, followed by a softmax operation to compute the corresponding likelihoods. This process ultimately yields a soft-labeled output suitable for downstream tasks.
  } 
  \label{fig:workflow}\vspace{-0.2in}
\end{figure}
\paragraph{Dataset Renovation via Missing Label Imputation}~
Motivated by these observations, we propose a dataset renovation framework \method for correcting noisy-label errors and addressing multi-label omissions in widely-used public image-classification test sets. Recent advances in Vision-Language Models (VLMs) offer promising solutions for this task, given their superior image recognition capabilities and robust performance in image-captioning tasks. For instance, BLIP \citep{li2022blip} leverages a pretrained VLM decoder to synthesize high-quality training data, illustrating the potential of utilizing pretrained VLMs' prior knowledge. Inspired by this insight, we introduce a comprehensive pipeline (illustrated in Figure \ref{fig:workflow}) that employs multiple state-of-the-art VLMs to renovate diverse test benchmarks and aggregates their outputs via a weighted voting ensembling strategy. By combining annotations derived from VLMs with targeted human judgments, our approach effectively reduces label noise and generates complementary labels in a reliable and cost-effective manner.

Our contributions include:
\squishlist
    \item \textbf{Missing Labels in Benchmarks:}  We highlight the critical yet overlooked issue of missing labels in widely-used image classification test sets, where images often contain multiple valid classes.
    \item \textbf{Systematic Analysis of VLMs:}  We investigate performances and limitations of VLMs in zero-shot image classification, revealing their strengths and weaknesses.
    \item \textbf{Revealing New Observations:}  We identify potential failure modes in VLM predictions, including hallucination, repetition, and reluctance to abstain, especially in fine-grained or large label sets. Missing labels often reflect uncertainty or true multi-object presence, where semantic confusion arises in dense label spaces. 
    \item \textbf{Test sets Renovation with Ensembling:}  Building on these observations, we propose \method, leveraging multiple VLMs to renovate six test benchmarks, aggregating predictions via an ensembling strategy to provide accurate soft labels and relevant annotations. \method shows strong alignment with human judgment in most datasets. 
\squishend

\section{Related Work}
\vspace{-0.05in}
\textbf{Label Noise Detection \& Missing Label Imputation}~
Noisy and missing labels remain major obstacles in image classification. The most reliable remedy, manual re‑annotation, scales poorly to today’s large datasets \citep{chang2017revolt,krivosheev2020detecting,wei2021learning,liu2024automatic}. Automated approaches typically involve label prediction or confusion matrix estimation \citep{goldberger2017training,hendrycks2018using,yu2019does,wei2020combating, tanno2019learning, chen2019understanding,xia2019anchor,kumar2021constrained}, yet they often underperform in the real-world label noise \cite{wang2019symmetric,amid2019robust,wang2021policy,ma2020normalized,liu2020peer,cheng2020learning,wei2021open,liu2025human}. Recent advances such as CLIP \citep{radford2021learning} have enabled zero-shot prediction techniques \citep{zhu2023unmasking, wei2024vision}, while the Confident Learning framework (CL) provides a principled way to identify noisy labels and model uncertainty, particularly under class imbalance \citep{northcutt2021confident, northcutt2021pervasive,wei2023fairness,wei2023distributionally}. Beyond noise, the issue of missing labels is particularly pronounced in multi-label settings, where exhaustive annotation is prohibitively expensive. To mitigate this, Durand et al.\ \citep{durand2019learning} proposed a partial binary cross-entropy loss and a curriculum learning strategy that iteratively infers missing labels using model feedback. Ben-Cohen et al.\ \citep{ben2022multi} introduced a temporary model to estimate the distribution of unobserved labels, coupled with an asymmetric loss that emphasizes known labels. Ma et al.\ \citep{ma2022label} preserved label semantics via contrastive embedding, while Zhang et al.\ \citep{zhang2023learning} proposed a unified framework that combines bootstrapped label correction, multi-focal loss, and balanced training. 

\textbf{Vision-Language Models for Classification Tasks}~
The emergence of CLIP \citep{radford2021learning} demonstrated that pretrained VLMs can leverage label semantics for zero-shot image classification. ALIGN \citep{jia2021scaling} scaled this approach using over a billion noisy image-text pairs, achieving strong performance on image-text retrieval benchmarks. Subsequent methods enhanced alignment and fusion \citep{li2021align,yao2021filip}, enabled multi-task learning and efficient fine-tuning (e.g., iCLIP \citep{wei2023iclip}, LiT \citep{zhai2022lit}), and improved classification via prompt optimization (e.g., CoCoOp \citep{zhou2022conditional}). Recent efforts have focused on few-shot generalization and instruction-following capabilities. Flamingo \citep{alayrac2022flamingo} and OpenFlamingo \citep{awadalla2023openflamingo} introduced cross-modal attention for strong performance across vision-language tasks. BLIP-2 \citep{li2023blip2} further advanced VLMs by connecting frozen image encoders with large language models through a query transformer. Instruction-tuned models like InstructBLIP \citep{dai2023instructblip} improved alignment with natural language prompts, enhancing robustness in open-ended settings. These developments underscore the potential of VLMs for open-vocabulary classification with minimal supervision.

\section{Are Vision-Language Models Good Label Predictors?}

In this section, we evaluate four state-of-the-art VLMs: BLIP~\citep{li2022blip}, Janus-Pro-7B~\citep{wu2024janus}, Qwen-VL-Plus~\citep{bai2023qwen}, and LLaVA-13B~\citep{liu2023llava}, for their effectiveness in detecting noisy and missing labels in the CIFAR-10 and CIFAR-100 datasets. We introduce the renovation prompt used for this evaluation as well as the experimental setup, and summarize their impacts on the performance of VLMs. For each model, we outline its capabilities, results and provide a brief exploratory analysis of the findings.
All models are evaluated under consistent settings outlined in Table~\ref{tab:Renovation setting}.
\subsection{Prompting Choices}
Prompts are the significant factor of VLM's performance. We systematically examine three types of prompting strategies for guiding VLMs in image classification tasks: Binary Questioning, Direct Multi-Label Selection, and Batched Multi-Label Selection. The three categories of prompts are illustrated in \cref{prompt_examples}. The following provides detailed descriptions of each prompting approach:
\squishlist
    \item \textbf{Binary Questioning:}  Asking 10 yes/no questions for CIFAR-10 (100 for CIFAR-100) per image. Each image requires 10 prompts (100 for CIFAR-100) to finish classification. This type of prompts leads to dramatic cost of running time in dataset with large amount of labels.
    \item \textbf{Direct Multi-Label Selection:}  Asking VLMs to choose all relevant labels from a fixed list of options (10 labels for CIFAR-10 and 100 for CIFAR-100). Each image requires a single prompt.
    \item \textbf{Batched Multi-Label Selection:}  We begin by partitioning the full set of labels into multiple batches, with each prompt containing only a single batch. VLMs are constrained to generate labels exclusively from the given prompt. Consequently, each image is processed with a number of prompts equal to the total number of label batches. 
\squishend

\begin{figure}[!htb]
  \centering
  \vspace{-0.05in}
  \includegraphics[width=1.0\linewidth]{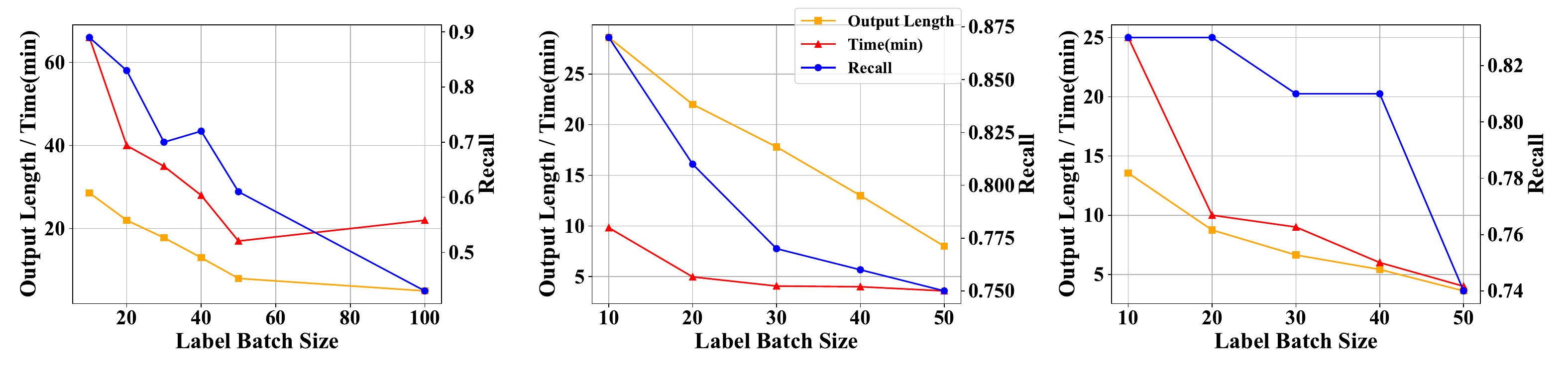}
  \vspace{-0.2in}
  \caption{Evaluation on different settings of prompt. Results shown from left to right are from \textbf{Janus}, \textbf{LLaVA}, \textbf{Qwen}, respectively. Results are evaluated on first 100 images of CIFAR-100 test data. To balance the running time and recall, our label batch size is set to be 20 accordingly.}
  \vspace{-0.1in}
  \label{fig: prompt_evaluation}
\end{figure}
To evaluate prompt design for VLMs, we assess Janus, LLaVA and Qwen on the first 100 images of the CIFAR-100 test set using recall, inference time, and output length. As shown in Figure~\ref{fig: prompt_evaluation}, larger label batch sizes reduce inference time by minimizing prompt iterations but tend to lower recall due to increased task complexity. A batch size of 20 offers an optimal trade-off, significantly improving efficiency while maintaining recall above 0.8, and is thus adopted for CIFAR-100 renovation. Beyond batch size, incorporating reasoning, image descriptions, and label shuffling enhances label quality by leveraging VLMs’ captioning capabilities and mitigating positional bias. Our final prompt design integrates all these elements. 

\subsection{Overall Experimental Setup}

Besides prompt setting, a few more parameters are required for running VLMs. As BLIP generates outputs through an Image-Text Matching (ITM) head, we extract the associated confidence scores and apply a threshold to filter out low-confidence labels, thereby enhancing labeling quality. Additionally, to prevent excessive label output and improve precision, we employ a top-$t$ filtering strategy where $t$ is determined based on the cardinality of the label set. To conclude, our setting includes \textbf{labels per prompt}, \textbf{threshold} and \textbf{top-}$t$ hyperparameters. The overall configuration for VLM-based dataset renovation is summarized in Table~\ref{tab:Renovation setting}. Details are shown in Appendix~\ref{app:experimental setting}.

\subsection{Observations and Performance }
Following the renovation process, we conduct a detailed analysis of the results, identifying several notable observations in the VLMs renovation results. Examples of our observations are presented and the general performance of each VLM on CIFAR-10 is also given. 
\textcolor{red}{\textbf{Observation 1:}}\textbf{ Failure in Fine-grained Class Prediction}~
Although instructed to select from a fixed set of labels, VLMs occasionally produce outputs that do not match any of the valid fine-grained candidates (e.g., sea snake), instead responding with more general terms (e.g., snake), which don't belong to candidates. This issue becomes more prevalent as the label set expands, potentially due to the insufficient grounding in the specified label space. 
\begin{tcolorbox}[colback=yellow!10, title=Observation 1: Failure in Fine-grained Class Prediction]
\vspace{-0.5em}
\begin{minipage}{0.15\textwidth}
    \includegraphics[width=\linewidth]{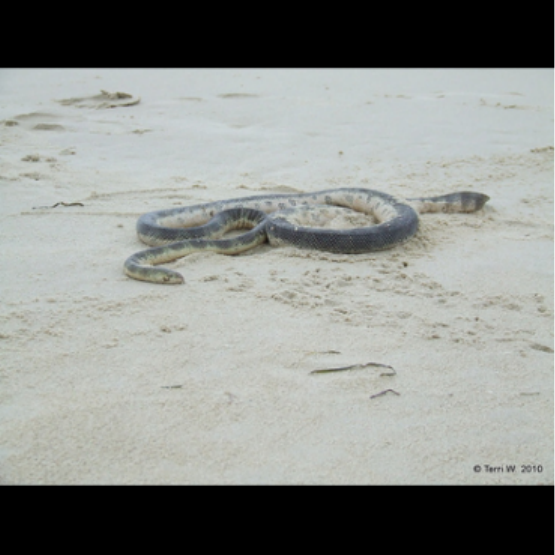}
\end{minipage}
\hspace{0.5em} 
\begin{minipage}{0.8\textwidth}
    \vspace{-0.2em}
    \small
    \textbf{Original Label:} ``sea snake''\\ 
    \textbf{Candidates:}[``sea snake'',``garter snake''...]\\
    \textbf{Response(by Janus):}\textit{ ``answer'': [``snake''], reason: ``The image shows a creature with distinct features such as scales covering its body, elongated shape without limbs, and forked tongue which is characteristic of snakes.''}
\end{minipage}
\vspace{-0.5em}
\end{tcolorbox}
\textcolor{red}{\textbf{Observation 2:}}\textbf{ Label Repetition and Stalling}~
Infinite repetition is occasionally observed when a large label batch (e.g., label batch size = 100) is provided. In some cases, identical labels are repeated multiple times within a single prediction, indicating that the model may become trapped in a repetitive loop. This behavior resembles a form of generation mode collapse, which diminishes both the diversity and informativeness of the model's outputs.
\begin{tcolorbox}[colback=yellow!10, title= Observation 2: Label Repetition and Stalling]
\begin{minipage}{0.15\textwidth}
\vspace{-0.5em}
    \includegraphics[width=\linewidth]{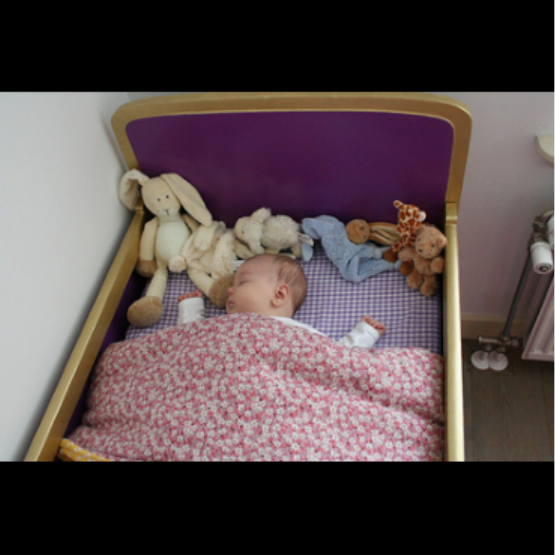} 
\end{minipage}
\hfill
\begin{minipage}{0.8\textwidth}
    \textbf{Original Label:} ``cradle'' \\
    \textbf{Response(by Janus):}\textit{ ``answer'': [``baby'', ``baby blanket'', ``baby toys'', ``baby crib'', ``baby sleeping bed with toys and blanket'',``baby sleeping bed with toys and blanket and crib and toys and toys and toys and toys...'']}
\end{minipage}
\vspace{-0.5em}
\end{tcolorbox}
\textcolor{red}{\textbf{Observation 3:}}\textbf{ Irrelevant Associative Reasoning}~
Even when the model correctly identifies the primary object in an image, it may still generate additional labels based on imaginative or weakly related associations. For instance, a model might correctly recognize a ``oak tree'' but subsequently generate labels such as ``apple'' despite their absence or irrelevance to the visual content.
\begin{tcolorbox}[colback=yellow!10, title=Observation 3: Irrelevant Associative Reasoning]
\vspace{-0.5em}
\begin{minipage}{0.15\textwidth}
    \includegraphics[width=\linewidth]{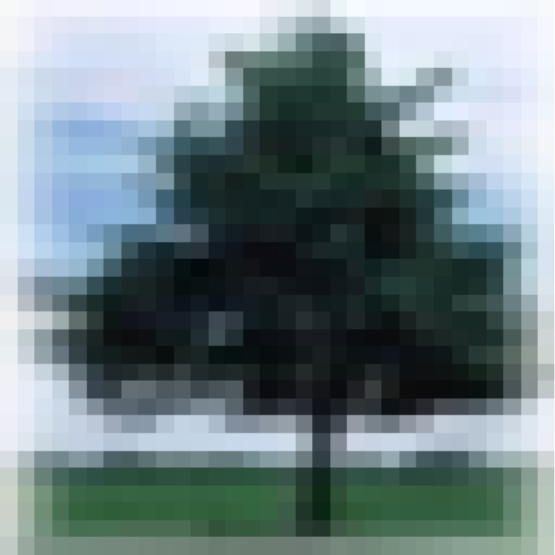} 
\end{minipage}
\hfill
\begin{minipage}{0.8\textwidth}
    \textbf{Original Label:} ``oak tree'' \\
    \textbf{Response(by LLaVA):}\textit{ ``answer'': [``apple''], ``reason'': ``The tree has a distinct shape and is surrounded by a grassy area, which is reminiscent of the natural environment where apples grow.''}
\end{minipage}
\vspace{-0.5em}
\end{tcolorbox}

\textcolor{red}{\textbf{Observation 4:}}\textbf{ Refusal of Null Response}~
In the context of batched multi-label selection, VLMs often fail to return ``None'' even when none of the candidate labels are relevant to the given image. This behavior undermines the precision of the model's responses and may stem from their training as cooperative assistants, which predisposes them to avoid issuing negative or null outputs.
\begin{tcolorbox}[colback=yellow!10, title= Observation 4: Refusal of Null Response]
\vspace{-0.5em}
\begin{minipage}{0.15\textwidth}
    \includegraphics[width=\linewidth]{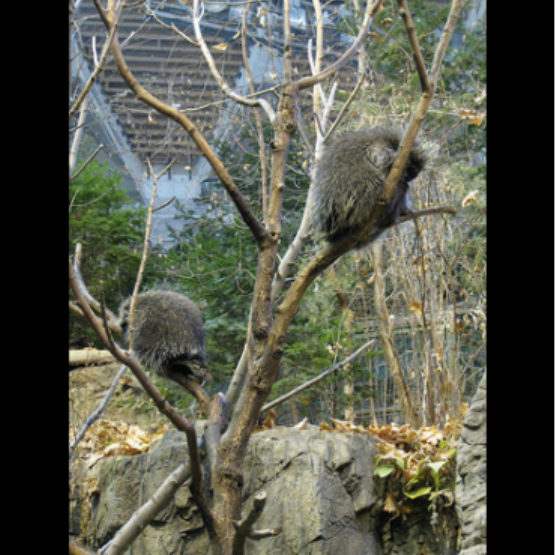} 
\end{minipage}
\hfill
\begin{minipage}{0.8\textwidth}
    \textbf{Original Label:} ``porcupine''\\
    \textbf{Candidate:}[``reel'',``radiator'',``sunglass'',``car mirror'',``tusks'']\\
    \textbf{Response(by Qwen):}\textit{ ``answer'': [``reel''],
    reason: ``The objects appear to resemble porcupines or similar animals perched on tree branches.''}
\end{minipage}
\vspace{-0.5em}
\end{tcolorbox}

\textbf{Performance across VLMs on CIFAR-10}~
\begin{figure}[!htb]
  \centering
  \vspace{-0.15in}
\includegraphics[width=1.0\textwidth]
  {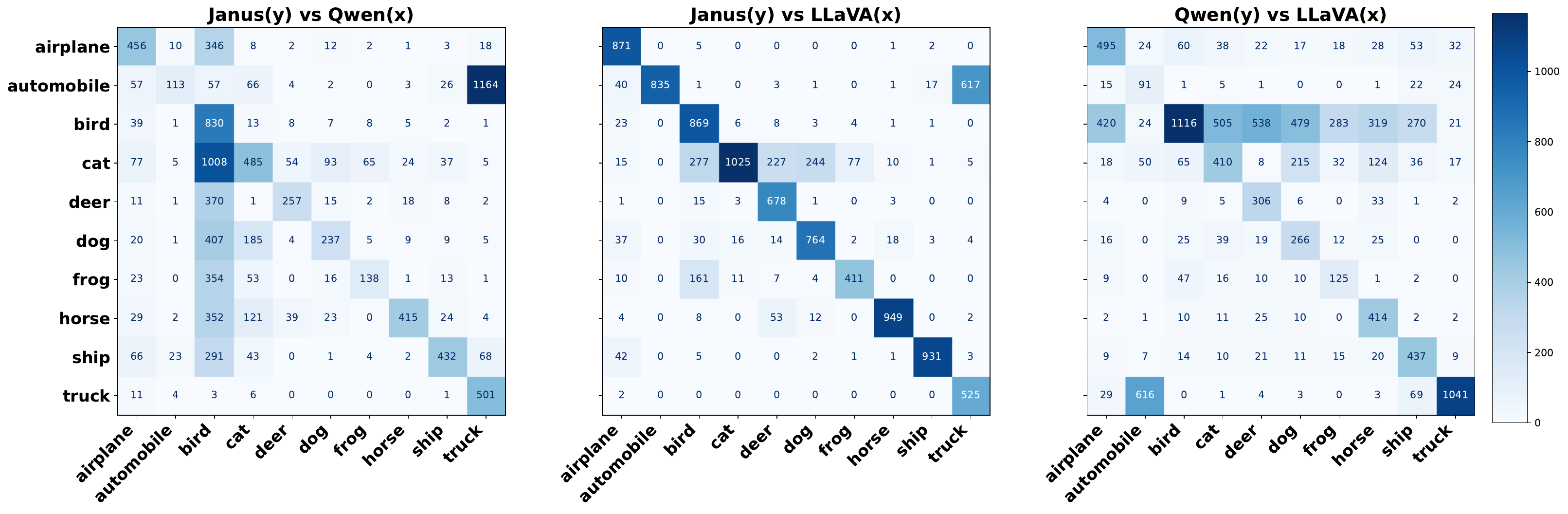}
  \vspace{-0.2in}
  \caption{Models Pairs Comparison. These three sub-figures illustrate confusion matrix of renovation results between VLMs: Janus/Qwen, Janus/LLaVA, Qwen/LLaVA respectively.}  \vspace{-0.15in}
  \label{fig:pairs_compare}
\end{figure}
To better understand the labeling behavior of vision language models on CIFAR-10, we analyzed both the agreement patterns between models and the frequency distribution of predicted classes. To locate biases and assess intermodel consistency, we constructed pairwise confusion matrices (Figure~\ref{fig:pairs_compare}). The Janus–LLaVA matrix exhibits strong alignment, especially for visually distinct classes such as airplane, bird, and truck, where predictions closely match across models. However, Qwen–LLaVA reveals pronounced divergence: Qwen heavily over-predicts bird, cat, and dog, leading to inflated confusion with semantically or visually related categories like deer and horse. This is further confirmed by the Janus–Qwen matrix, where Qwen’s tendency to misclassify ambiguous or fine-grained categories results in substantial off-diagonal errors. These discrepancies indicate that while models share consensus on coarse or high-contrast categories, interpretive biases and over-generalizations persist in handling nuanced or overlapping visual features, causing varied predictions in finer-grained classification tasks.
\begin{wrapfigure}[11]{r}{0.47\textwidth}
  \centering
  \vspace{-0.16in}
\includegraphics[width=0.95\linewidth]{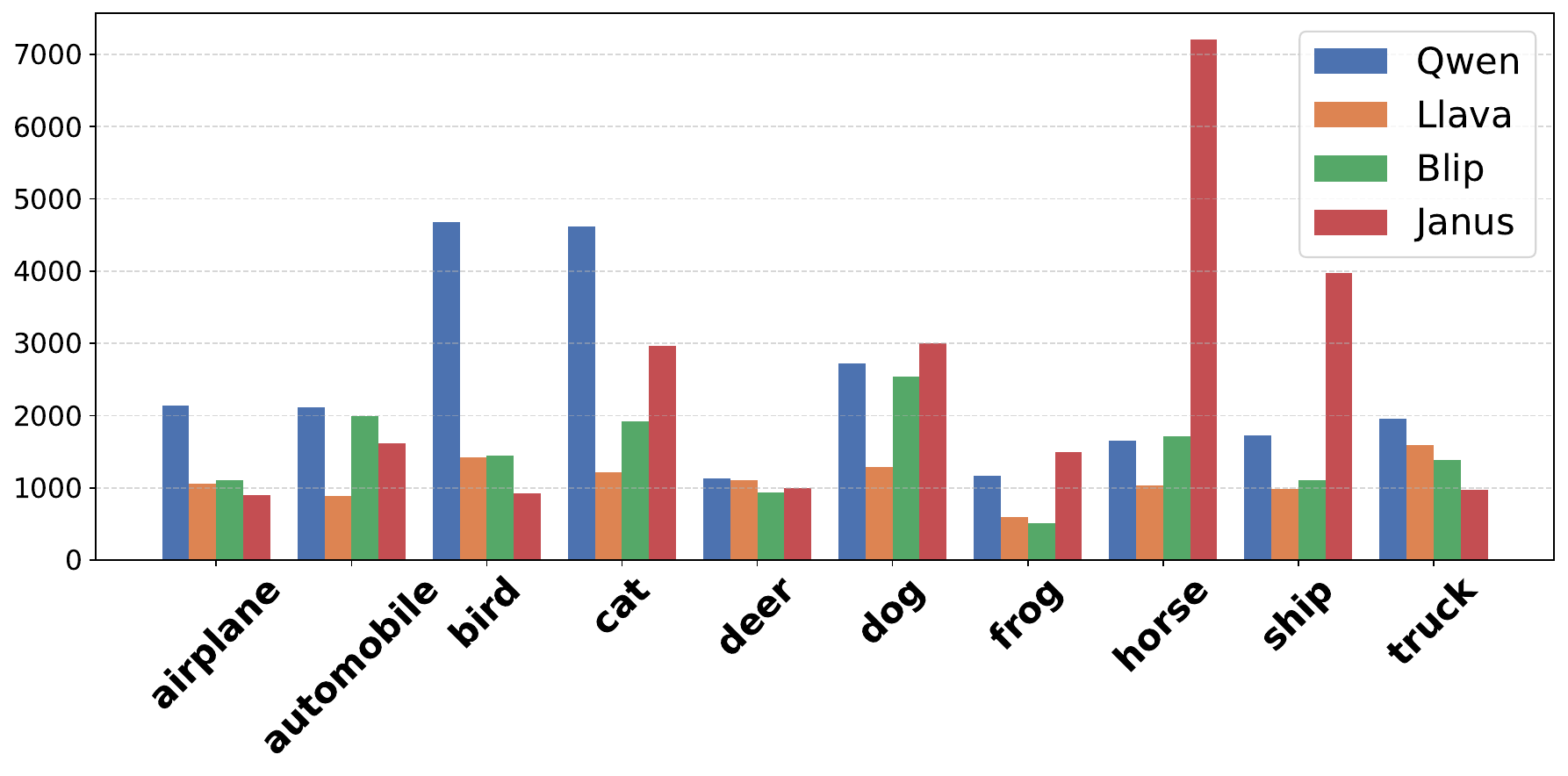}
  \vspace{-0.1in}
  \caption{Count of VLMs' output on CIFAR-10.}
  \vspace{-5.1in}
  \label{fig:model_performance}
\end{wrapfigure}
The label distribution is presented in Figure~\ref{fig:model_performance} illustrating all four models exhibit predictions across all ten CIFAR-10 classes. The figure reveals a notable observation: VLMs exhibit biased prediction tendencies across different classes. In particular, Qwen shows a strong preference for the label bird and cat, frequently assigning them more than other categories. Janus tends to significantly overpredict horse, while LLaVA and BLIP display more balanced label distributions. No significant bias found in these two models. We believe that the presence of these biases may be attributed to hallucinations caused by prior knowledge or learning objectives used during model development. In particular, imbalanced label distributions or domain-specific image-text pairs could reinforce model preferences for certain categories.

\section{Equipping VLM with Label-Noise Curation}
In addition to VLMs, we also incorporate several specialized label diagnosis methods, including Cleanlab and Docta, to support and enhance the final label aggregation process. Cleanlab \citep{northcutt2021confident} is a model-agnostic framework that identifies and corrects noisy labels by estimating the joint distribution of noisy and true labels using the Confident Learning paradigm. Docta \citep{zhu2023unmasking} systematically audits dataset credibility by estimating label noise and ranking instances based on the alignment between observed and inferred labels without requiring ground-truth annotations.

For the results obtained from VLMs and related methods, our aggregation procedure consists of the following processes: 


\subsection{Constructing Pseudo Ground-Truth Labels via Model Voting}

For each dataset $X$, we estimate the accuracy of each method using the first 100 images. Specifically, we collect predictions from $m$ different models $\{\tilde{y}_1, \tilde{y}_2, \dots, \tilde{y}_m\}$, where each $\tilde{y}_i = [\tilde{y}_i^{(1)}, \tilde{y}_i^{(2)}, \dots, \tilde{y}_i^{(n)}]$ denotes the predicted label set by model $i$ on the first $n=100$ images.

For each image $x^{(j)}$, we apply a voting strategy across the $m$ models to generate pseudo ground-truth labels. Specifically, the vote count for a candidate label $c \in \mathcal{C}$ is defined as:
\[
\text{vote}^{(j)}(c) = \sum_{i=1}^{m} \mathbb{I}\left[c \in \tilde{y}_i^{(j)}\right],
\]
where $\mathbb{I}[\cdot]$ is the indicator function, equal to 1 if model $i$ predicts label $c$ for image $x^{(j)}$, and 0 otherwise. The aggregated pseudo ground-truth label set for image $x^{(j)}$ is defined as:
\[
\tilde{y}_{\text{ground\_truth}}^{(j)} = \left\{c \in \mathcal{C} \mid \text{vote}^{(j)}(c) \geq k \right\},
\]
where \( k \) is the minimum number of votes required for label \( c \) to be included. The ground truth labels \( y_{\text{ground\_truth}} \), manually verified, are used to assess the model's estimated accuracy.

\subsection{Model Expertise (Accuracy) Estimation}

Given the pseudo ground-truth labels $\tilde{y}_{\text{ground\_truth}}$, we assess the  estimated accuracy of each model on this subset. To account for the tendency of VLMs to produce overly broad predictions, we introduce a regularization term that penalizes such behavior during evaluation.

Let $\tilde{y}_{\text{model}}^{(j)}$ denote the predicted label set of a model for image $x^{(j)}$, and let $\tilde{y}_{\text{ground\_truth}}^{(j)}$ be the corresponding pseudo ground-truth label set. We compute model estimated accuracy as:
\[
\text{est\_acc}:= \left( \frac{\sum_{j=1}^{n} \left| \tilde{y}_{\text{model}}^{(j)} \cap \tilde{y}_{\text{ground\_truth}}^{(j)} \right|}{\sum_{j=1}^{n} \left| \tilde{y}_{\text{ground\_truth}}^{(j)} \right|} \right) \cdot \left(1 - \frac{\sum_{j=1}^{n} \left| \tilde{y}_{\text{model}}^{(j)} \right|}{n \cdot |\mathcal{C}|} \right).
\]
Here, $|\cdot|$ denotes set cardinality, $n$ is the number of images (typically 100), and $\mathcal{C}$ is the complete set of candidate labels. The second term penalizes models that produce excessive predictions, thus reducing the impact of random guessing. This adjustment encourages more precise outputs and improves robustness in accuracy estimation. The estimated accuracies of different models across datasets are summarized in Table~\ref{tab:method_accuracy}. 
\vspace{-0.05in}
\begin{table}[h!]
\centering
\vspace{-0.06in}
\scalebox{0.9}{
\begin{tabular}{lcccccccc}
\toprule
\textbf{Dataset} & \textbf{BLIP} & \textbf{LLaVA} & \textbf{Janus} & \textbf{Qwen} & \textbf{DOCTA} & \textbf{Cleanlab} & \textbf{Origin} &\textbf{Full Score} \\
\midrule
CIFAR-10     & 0.777 & 0.815 & 0.676 & 0.650 & 0.973  & 0.928  & 0.975 & 5.794\\
CIFAR-100    & 0.750 & 0.812 & 0.650 & 0.774 & 0.839  & 0.671  & 0.856 & 4.730\\
Caltech256   & 0.927 & 0.843 & 0.784 & 0.879 & --     & 0.808  & 0.935 & 5.176 \\
ImageNet 1K  & 0.858 & 0.807 & 0.490 & 0.793 & --     & 0.490  & 0.569 & 4.007 \\
QuickDraw    & 0.710 & 0.773 & 0.430 &  0.792 & --    & 0.181  & 0.210 & 3.096\\
MNIST        & 0.592 & 0.696 & 0.582 & 0.822 & 0.983  & 0.988  & 0.991 & 5.654\\
\bottomrule
\end{tabular}
}
\caption{Estimated accuracy results of different renovation methods across datasets. \textbf{Origin} represents the dataset's original label estimated accuracy evaluated on pseudo ground truth. \textbf{Full Score} is the sum of all the methods' accuracies, denoting the theoretical score for the most likely labels.}
\label{tab:method_accuracy}
\end{table}
\vspace{-0.2in}

Results reveal a negative correlation between label pool size and estimated accuracy, underscoring the impact of task complexity on model performance. VLMs achieve the highest accuracy on the renovated Caltech256 dataset, likely due to its higher-resolution images, and perform well on ImageNet-1K, where Cleanlab underperforms. Conversely, Cleanlab outperforms VLMs on MNIST, indicating that statistical or human-guided methods are more effective for structured data. These findings suggest a complementary relationship between VLMs and traditional renovation methods, contingent on dataset characteristics.

\subsection{Weighted Aggregation using Estimated Accuracy Scores}

Following the estimated accuracy, we perform prediction aggregation using the estimated accuracy scores $\{\text{Acc}_1, \dots, \text{Acc}_m\}$ as model weights, without applying softmax normalization. For each image $x^{(j)}$ and candidate label $c \in \mathcal{C}$, we compute a weighted support score as:
\[
\text{score}^{(j)}(c) = \sum_{i=1}^{m} \text{Acc}_i \cdot \mathbb{I}\left[c \in \tilde{y}_i^{(j)}\right],
\]
where $\mathbb{I}[\cdot]$ denotes the indicator function. This score quantifies the level of support for label $c$ across all models, adjusted by their estimated reliability.

\subsection{Post-Aggregation Filtering and Normalization}

To determine the final label set for each image $x^{(j)}$, we apply a two-step filtering and normalization procedure:
\squishlist
    \item \textbf{Thresholding and Top-$n$ Selection}  ~
    For each image $x^{(j)}$, we retain labels $c$ that satisfy both:
    (i) $\text{score}^{(j)}(c) \geq \tau$, where $\tau$ is a predefined threshold; and
    (ii) $c$ ranks among the top $n$ labels with the highest scores for $x^{(j)}$.
    Let $\mathcal{C}_{\text{filtered}}^{(j)}$ denote the resulting label set.

    \item \textbf{Softmax Normalization}  ~
    We normalize the scores within $\mathcal{C}_{\text{filtered}}^{(j)}$ via softmax: 
    \[
    p^{(j)}(c) = \frac{e^{\text{score}^{(j)}(c)}}{\sum_{c' \in \mathcal{C}_{\text{filtered}}^{(j)}} e^{\text{score}^{(j)}(c')}}, \quad c \in \mathcal{C}_{\text{filtered}}^{(j)}.
    \]
\squishend

\begin{table}[h!]
\centering
\scalebox{0.9}{
\begin{tabular}{lccccc}
\toprule
\textbf{Dataset} & \textbf{Image number} & \textbf{Noisy Label} & \textbf{Missing label} & \textbf{Threshold/Full score} & \textbf{Top-}$K$ \\
\midrule
CIFAR-10     & 10000 & 38 & 1325 & \textbf{0.900}/5.794 & \textbf{3}\\
CIFAR-100 & 10000 & 552 & 6083 & \textbf{0.830}/4.730 & \textbf{5} \\
Caltech256 & 30607 & 766 & 30187 & \textbf{0.037}/5.176 & \textbf{7} \\
Quickdraw & 2500 & 462 & 2500 & \textbf{0.018}/3.096 & \textbf{5} \\
Imagenet & 50000 & 6546 & 35147 & \textbf{0.300}/4.007 & \textbf{10}\\
Mnist & 10000 & 24 & 469 & \textbf{0.950}/5.654 & \textbf{3} \\
\bottomrule
\end{tabular}
}
\caption{\method results across datasets. Here, \textbf{Top-}$K$ denotes the maximum number of labels allowed per image in the dataset, and the \textbf{Threshold} refers to the minimum weighted support score required for a label to be included in the final result. The definition of \textbf{Full Score} has been introduced in Table~1. \textbf{Noisy Label} and \textbf{Missing Label} represent the number of images identified by \method as containing such issues under the specified configuration.}
\label{tab:agg_results}
\end{table}
\subsection{Observations}

\begin{figure}[htbp]
  \centering
  \vspace{-1em}
  \includegraphics[width=0.96\textwidth]{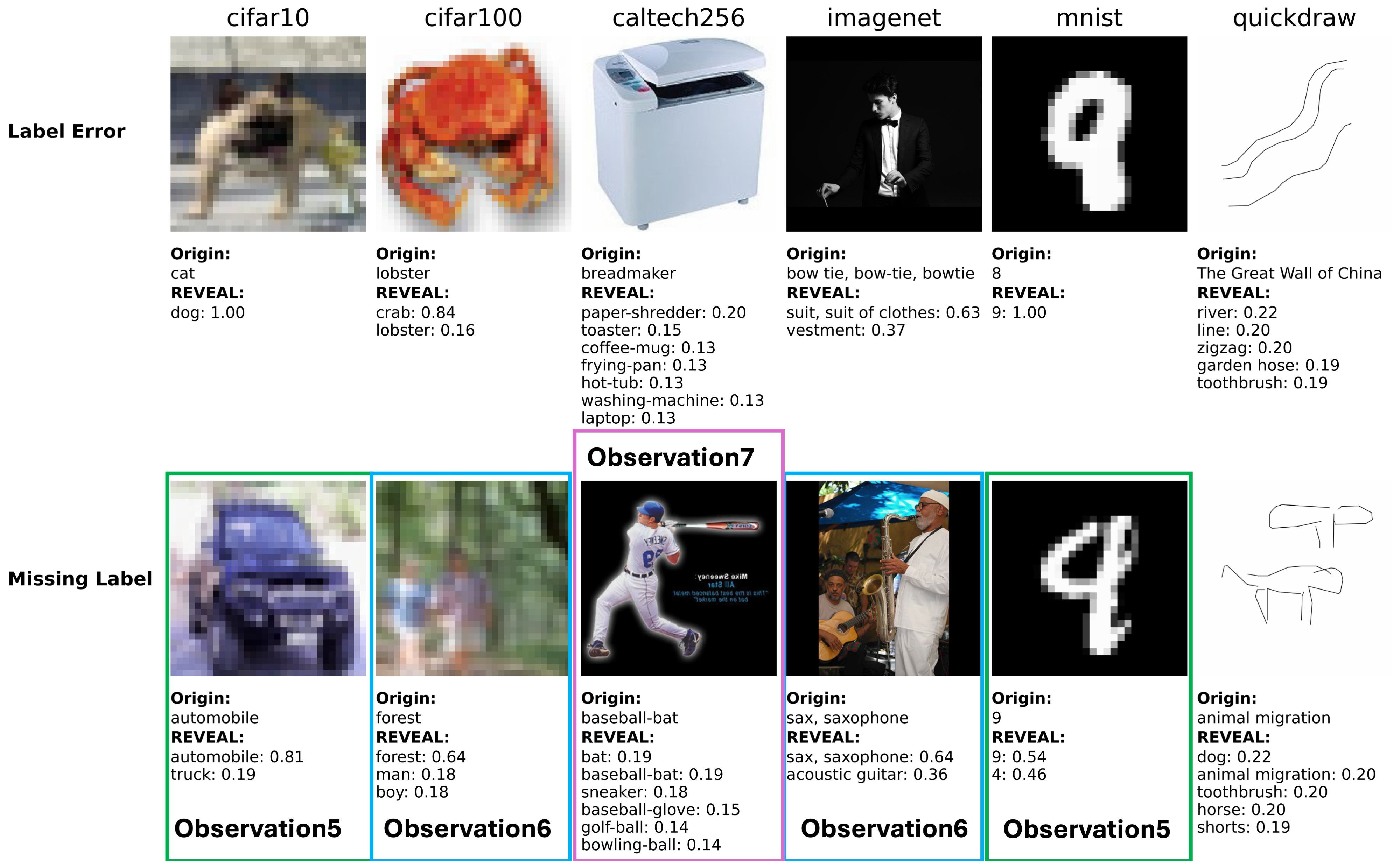}\vspace{-0.1in}
  \caption{Label correction visualization across datasets.}
\vspace{-1em}
  \label{fig:label_viz}
\end{figure}


\textcolor{red}{\textbf{Observation 5:}} \textbf{Missing labels are another possible way to describe the image content.} ~In datasets with a relatively small number of possible labels, such as CIFAR-10 and MNIST, when multiple missing labels are identified through renovation, this ``multiplicity'' often reflects uncertainty in label assignment rather than the actual co-occurrence of multiple objects. For instance, as illustrated in a missing-label example from MNIST in Figure~\ref{fig:label_viz}, VLMs estimate a 54\% probability that the digit is a 9 and a 46\% probability that it is a 4, rather than suggesting that both digits are simultaneously present in the image.

\textcolor{red}{\textbf{Observation 6:}}\textbf{ Missing labels also refer to elements present in the image but omitted during annotation.} ~In datasets with a large number of possible labels, such as CIFAR-100 (100 classes) and Caltech256 (257 classes), cases of multiple missing labels identified through renovation are more likely to indicate true multi-object presence. As illustrated in a missing-label example from CIFAR-100 in Figure~\ref{fig:label_viz}, VLMs predict that, in addition to the original label ``forest'', the image also contains ``man'' and ``boy'', suggesting genuine label omissions rather than probabilistic ambiguity.

\textcolor{red}{\textbf{Observation 7:}}\textbf{ When the number of label candidates is extremely large (e.g., in the case of ImageNet), VLMs tend to produce semantically related sets of labels.}~ In datasets with a large number of possible labels, such as Caltech256 and ImageNet, the occurrence of multiple missing labels during renovation may sometimes be attributed to semantic similarity between certain classes. As illustrated in an example from Caltech256 in Figure~\ref{fig:label_viz}, VLMs predict that the image could correspond to either ``baseball bat'' or ``bat''. While traditional deep learning classification models typically treat labels as discrete one-hot encodings, VLMs interpret label semantics in a continuous space, making them more prone to confusion between semantically similar classes.

\subsection{Comparison with Human-Annotated Renovation}

We further compare and analyze the renovation results produced by our VLMs and other methods against human annotations collected via MTurk~\citep{northcutt2021labelerrors} across the datasets. We calculated the agreement rates between the renovation results of our four VLM-based methods and the aggregated final results with the MTurk-evaluated datasets (as shown in Table~\ref{tab:mturk_evaluation}).
\begin{table}[h!]
\centering
\scalebox{0.86}{
\begin{tabular}{l|ccccc}
\toprule
\textbf{Dataset} & \textbf{BLIP} & \textbf{LLaVA} & \textbf{Janus} & \textbf{Qwen} & \method \\
\midrule
CIFAR-10     & 0.806 & 0.785 & 0.818 & 0.704 & \textbf{\cellcolor{green!30}0.976} \\
CIFAR-100    & 0.661 & 0.755 & 0.556 & 0.668 & \textbf{\cellcolor{green!30}0.880} \\
Caltech256   & 0.850 & 0.774 & 0.764 & 0.836 & \textbf{\cellcolor{green!30}0.946} \\
ImageNet 1K  & \textbf{\cellcolor{green!30}0.664} & 0.560 & 0.306 & 0.402 & 0.569 \\
QuickDraw    & 0.369 & 0.421 & 0.260 & 0.467 & \textbf{\cellcolor{green!30}0.630} \\
MNIST        & 0.435 & 0.518 & 0.365 & 0.506 & \textbf{\cellcolor{green!30}0.776} \\
\bottomrule
\end{tabular}
}
\caption{Agreement rates between \method and MTurk annotations across datasets. Individual VLMs and the aggregated results (\method) are evaluated against MTurk labels. \textbf{\colorbox{green!30}{Highlighted}} cells indicate the highest agreement per dataset.}
\label{tab:mturk_evaluation}
\end{table}

\begin{figure}[htbp]
  \centering
  \vspace{-2.3em}  \includegraphics[width=0.9\textwidth]{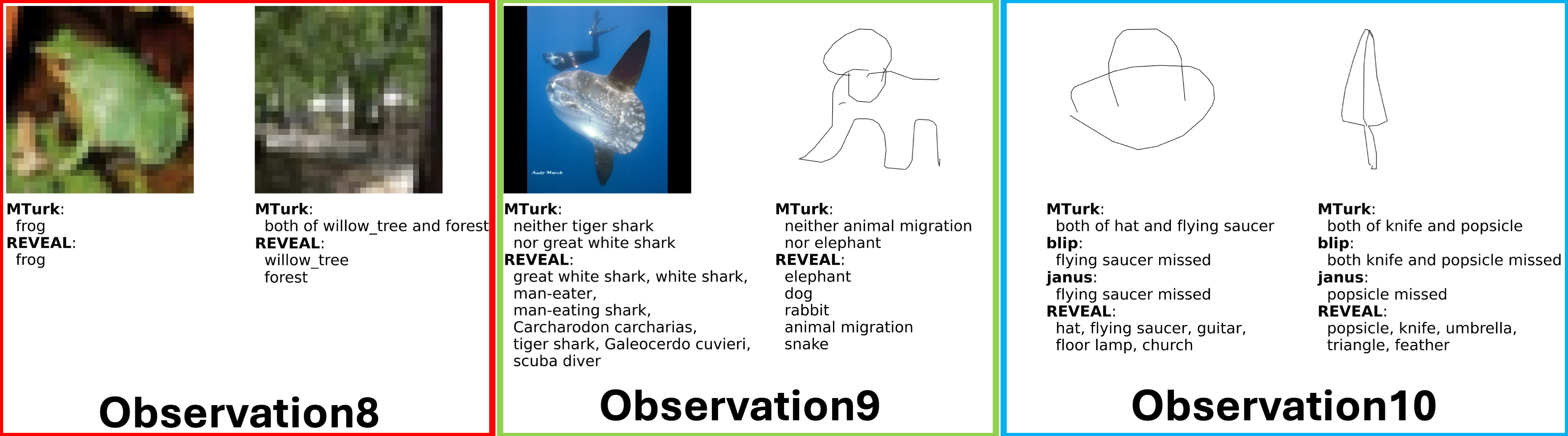}  \caption{Examples comparing \method results with human annotations and single VLM results.}
  \label{fig:vsmturk}
  \vspace{-0.9em}
\end{figure}
With respect to the comparison against MTurk human annotations, we give:

\textcolor{red}{\textbf{Observation 8:}} \textbf{High Alignment with Human Annotations} Across most datasets, \method method demonstrates a high degree of consistency with MTurk human annotations.

\textcolor{red}{\textbf{Observation 9:}} \textbf{Discrepancies Due to Mturk Constraints} On the ImageNet and QuickDraw datasets, \method exhibits substantial discrepancies compared to the MTurk annotations, which may arise from limitations in the MTurk evaluation protocol. In particular, annotators are typically limited to validating only two candidate labels per image. For datasets like ImageNet and QuickDraw, which have a large space of plausible labels, this constraint fails to adequately capture missing label issues, thereby reducing the reliability of the MTurk annotations on these datasets.

\textcolor{red}{\textbf{Observation 10:}} \textbf{\method Outperforms Individual VLM} Across most datasets, \method demonstrates superior alignment with human judgment compared to individual VLMs.




\section{Conclusion}

 We address a critical yet underexplored issue in image classification: the presence of noisy and missing labels in widely-used benchmark datasets. We propose a unified framework, \method, that renovates test sets by combining label noise detection and missing label imputation using VLMs. Through model agreement analysis, expertise estimation, and ensembling, we construct high-quality pseudo ground-truths that better reflect image content. Across six datasets: CIFAR-10, CIFAR-100, ImageNet, Caltech256, QuickDraw, and MNIST, alongside comparison with human annotations (MTurk), we find noisy labels and missing labels to be pervasive, and we provide a systematic analysis via \textbf{{\color{red}10 observations}}. Our method effectively reveals missing labels, providing \textbf{soft-labeled} outputs that exhibit a \textbf{high degree of alignment with human judgments}. 
 Our work offers a model-centric alternative for benchmark improvement, and we hope it inspires future efforts in multi-label evaluation, open-vocabulary testing, and human-in-the-loop verification. 

\newpage

\bibliographystyle{plain}
\bibliography{noise_learning,myref,old_noise_learning}


\newpage
\appendix


\section{Broader Impacts and Limitations}
In this section, we list some of the broader impacts as well as the limitations of \method. 
\subsection{Border Impacts}
\squishlist
    \item \textbf{Vision-language Dataset Renovation}  
    
    \method is not limited to image classification and can be extended to other multi-modal tasks that require strong visual recognition capabilities. Image captioning is a particularly suitable application, as VLMs are explicitly trained to generate descriptive textual outputs from images. However, many existing image captioning datasets suffer from noise and sub-optimal quality, largely due to their collection from uncontrolled web sources. For instance, captions relying solely on proper nouns of tourist attractions fail to convey the actual visual content of the scene. This underscores the need for effective data curation strategies. By applying our weighted voting ensemble approach, \method can produce more accurate, abundant and fine-grained image descriptions, leading to the development of higher-quality benchmarks for training and evaluating multi-modal models.
    
    \item \textbf{Biased Dataset Curation via Ensembling}  
    
    The aggregation design in \method enables the correction of biases inherent in origin datasets or individual curation methods. As VLMs are prone to hallucinations and biased output due to their strong priors, employing a weighted voting scheme based on estimated accuracies helps mitigate noise and erroneous outputs from any single method. This corrective effect is empirically observed in our renovated results, demonstrating the robustness of the ensemble approach.

    \item \textbf{Soft Labels and Extensions}  

    \method effectively identifies missing labels and provides soft-labeled outputs accompanied by likelihood estimates, facilitating its generalization to diverse classification tasks. Compared to traditional one-hot encoding approaches, soft-labeling offers a more nuanced representation that integrates both linguistic and visual semantic information, thereby aligning predictions more closely with human perception and enhancing explainability. Such representations are particularly beneficial for semantic analysis tasks across both natural language processing and computer vision domains. For example, soft-labels can naturally extend to object detection scenarios and sentinel analysis, where texts or image segments may correspond to multiple labels with associated likelihoods, improving prediction quality and semantic interpretability.

    \item \textbf{Understanding Perceptual Gaps Between Models and Humans}
    
    Beyond the renovation of test sets, \method also provides a valuable lens into human and AI agreement. By comparing its soft labeled outputs against human annotations, the framework can help identify systematic divergences between model reasoning and human judgment, especially in ambiguous or low-quality images. This opens a promising direction for studying the cognitive boundaries of alignment between vision language models and human perception, which is essential for building more transparent, trustworthy, and human-compatible AI systems.

\squishend

\subsection{Limitations}
\squishlist



    \item \textbf{Human Disagreement on Ambiguous Images} 
    
    Although we compare \method outputs against human annotations, some images in the datasets are inherently low-quality or visually ambiguous. This leads to variability in human judgment, making it difficult to establish a definitive ground truth and complicating the evaluation of model performance.
    \item \textbf{Limitations of Fixed Label Sets in Real Conditions} 
    
    In practical scenarios, images may contain objects or concepts that are not included in the predefined label set. As a result, even semantically reasonable predictions can be considered incorrect during evaluation, making it difficult to fully capture model performance in some conditions.
\squishend

\section{Prompt Evaluation Supplementary Results}
\label{app:prompt_evaluation_result}
Beyond label batch size, our experiments also reveal that prompting models to provide a reason for each decision yields higher recall at a fixed batch size. Moreover, leveraging the strong image captioning capabilities of VLMs, by first prompting for a description and then guiding label selection based on that description, further enhances labeling quality. In addition, we observe that VLMs occasionally generate labels that fall outside the predefined label set. To mitigate the potential bias introduced by an alphabetic ordering implicitly learned by the VLMs, the most effective strategy is to shuffle the label list prior to dividing it into batches. Accordingly, our final prompt design incorporates reasoning, image description, and shuffling labels.

{\small
\begin{tcolorbox}[colback=blue!5!white, colframe=gray!70!black, title=Prompt Examples]
\label{prompt_examples}
\textbf{\textless User Prompt\textgreater}: \textless Image placeholder \textgreater 
 Please follow the instructions with no exceptions. 

\vspace{0.5em}
\begin{tcolorbox}[colback=yellow!10, title=Binary Questioning]
\textit{Is the main object of this image an <label name>? All the answer should be in json format \{`answer':`Yes or No',`reason':`reason of the answers'\}} 
\end{tcolorbox}

\begin{tcolorbox}[colback=yellow!10, title=Direct Multi-Label Selection]
\textit{You are Given an image and answer all the labels that appear the image from the following options:(<label candidate list>) There may be multiple labels in a single image, please answer at most 3 possible labels and separate them with ‘,’. If there is no label appearing in the image, please answer None. Please provide the reason of replying these label and review your answer. Please think step by step and provide similar characteristic in details between the label you choose and the image.All the answer should be in format \{`answer':`your answer 1,your answer 2',`reason':`reason of the answers'\}}
\end{tcolorbox}

\begin{tcolorbox}[colback=yellow!10, title=Batched Multi-Label Selection]
\textit{
    Forget you previous answer. Describe the image and choose labels 
    from the candidates:(<label candidate list>) that you think are in
    the image. Remember your answer should only contain labels from the given candidates! 
    If you think none of them are in the image, please reply None. 
    Please provide a short reason for your choice
    Before you make the final response, carefully review if your answer ONLY contains labels in the candidates. Your answer should be a json dict:
    \{`answer': [your answer list], `description': image description, `reason':your reason for choosing them\}. Please don't reply in other formats.
    } 
\end{tcolorbox}
\end{tcolorbox}}

The evaluation results of label batch size using Qwen, Janus and LLava are presented in Table~\ref{tab:Qwen_prompt_test}, Table~\ref{tab:Janus_prompt_test}, Table~\ref{tab:LLaVA_prompt_test} respectively. 

\begin{table}[h!]
\centering
\scalebox{0.9}{
\begin{tabular}{lcccc}
\toprule
\textbf{Label Batch Size} & \textbf{Model}& \textbf{Recall} & \textbf{Output Length} &  \textbf{Time(min)} \\
\midrule
10   & Qwen & 0.83 & 13.56 & 25 \\
20   & Qwen & 0.83 & 8.75 & 10 \\
30   & Qwen & 0.81 & 6.64 & 9\\
40   & Qwen & 0.81 & 5.42 & 6\\
50   & Qwen & 0.74 & 3.61 & 4\\
\bottomrule
\end{tabular}
}
\caption{Prompt batch size evaluation results on Qwen}
\label{tab:Qwen_prompt_test}
\end{table}

\begin{table}[h!]
\centering
\scalebox{0.9}{
\begin{tabular}{lcccc}
\toprule
\textbf{Label Batch Size} & \textbf{Model}& \textbf{Recall} & \textbf{Output Length} &  \textbf{Time(min)} \\
\midrule
10   & Janus & 0.89 & 28.6 & 66 \\
20   & Janus & 0.83 & 22 & 40 \\
30   & Janus & 0.70 & 17.8 & 35\\
40   & Janus & 0.72 & 13 & 28\\
50   & Janus & 0.61 & 8 & 17\\
100  & Janus & 0.43 & 5 & 22\\
\bottomrule
\end{tabular}
}
\caption{Prompt batch size evaluation results on Janus}
\label{tab:Janus_prompt_test}
\end{table}

\begin{table}[h!]
\centering
\scalebox{0.9}{
\begin{tabular}{lcccc}
\toprule
\textbf{Label Batch Size} & \textbf{Model}& \textbf{Recall} & \textbf{Output Length} &  \textbf{Time(min)} \\
\midrule
10   & LLaVA & 0.87 & 28.6 & 9.82 \\
20   & LLaVA & 0.81 & 22 & 4.97 \\
30   & LLaVA & 0.77 & 17.8 & 4.07\\
40   & LLaVA & 0.76 & 13 & 4.00\\
50   & LLaVA & 0.75 & 8 & 3.59\\
\bottomrule
\end{tabular}
}
\caption{Prompt batch size evaluation results on LLaVA}
\label{tab:LLaVA_prompt_test}
\end{table}

\section{Experimental Setting}
\label{app:experimental setting}
The experimental setting for individual renovation methods are concluded in Table~\ref{tab:Renovation setting}. For renovation, we deploy each of three VLMs (BLIP\citep{li2022blip}, LLaVA\citep{liu2023llava}, and Janus\citep{wu2024janus}) via 16 machines with NVIDIA A800 GPU (80GB memory). Our batch size is set to 32 for BLIP as well as LLaVA and 1 for Janus. Renovation of Qwen is carried out by API access and therefore is executed on a CPU (13th Gen Intel(R) Core(TM) i7-13620H (16CPUs)) with 5 processes for parallel processing. Due to limited computational resources, Docta\citep{zhu2023unmasking} can only perform diagnostics on datasets with a relatively small number of label candidates (e.g., CIFAR-10, CIFAR-100, and MNIST). All experiments were conducted on a single machine equipped with one NVIDIA L20 GPU (48GB memory). For detailed configurations, please refer to the official Docta repository: https://github.com/Docta-ai/docta.

\begin{table}[h!]
\centering
\begin{tabular}{lccc}
\toprule
\textbf{Dataset} & \textbf{Threshold(for BLIP)} & \textbf{Top-$\alpha$(for BLIP)} &  \textbf{label/prompt(for other VLMs)} \\
\midrule
CIFAR-10    & 0.15 & 3 & 10 \\
CIFAR-100   & 0.015 & 5 & 20 \\
Caltech256  & 0.006 & 5 & 50\\
ImageNet 1K    & 0.00015 & 20 & 67\\
QuickDraw& 0.004 & 5 & 60\\
Mnist       & 0.15 & 3 & 10 \\
\bottomrule
\end{tabular}
\caption{Renovation settings across datasets.}
\label{tab:Renovation setting}
\end{table}

\section{Introduction to Methods}
\paragraph{BLIP.} BLIP (Bootstrapping Language-Image Pre-training) \citep{li2022blip} is a VLM that includes an Image-Text Matching (ITM) function, which measures the compatibility between an image and a given text prompt. To adapt BLIP for multi-class classification, we compute matching scores between each image and all possible class labels (10 for CIFAR-10, 100 for CIFAR-100). The resulting scores are passed through a softmax transformation to yield a probability distribution over the candidate labels. A threshold is then applied to determine which labels are retained. We deploy \textbf{BLIP-2} locally. 

BLIP-2\citep{li2023blip2} distinguishes itself through a two-stage pre-training framework that bootstraps supervision from noisy image-text pairs. In the first stage, BLIP uses a captioning model to generate synthetic captions for images, filtering out low-quality pairs. In the second stage, it uses these refined pairs to train both image-text contrastive and matching objectives. This bootstrapped mechanism enables BLIP to learn more accurate alignment between modalities and improves its zero-shot classification performance.

\paragraph{Janus.} Janus \citep{wu2024janus}, developed by DeepSeek-AI, presents a unified multimodal framework that decouples visual encoding for better performance in both understanding and generation tasks, which makes it suitable for complex multimodal applications. We deploy \textbf{Janus-Pro-7B} locally.

What makes Janus-Pro unique is its dual-stream architecture, which explicitly separates visual understanding and generation capabilities\citep{chen2025januspro}. Unlike models that fuse modalities early, Janus maintains independent pathways for encoding image and text representations, allowing it to better preserve modality-specific features. This decoupled approach, paired with a shared cross-modal attention mechanism, results in high versatility across a wide range of tasks from classification to multimodal reasoning and instruction following.

\paragraph{Qwen.} Qwen-VL \citep{bai2023qwen} is part of Alibaba Cloud's Qwen series. Qwen-VL series excels in tasks requiring fine-grained visual understanding and localization, supporting multilingual interactions, and demonstrating strong performance in various vision-language benchmarks. We accessed \textbf{Qwen-VL-Plus} through the API as it is not ideal to be deployed locally.

Qwen-VL-Plus is notable for its rich grounding capability, which tightly couples objects in the image with their semantic descriptions. It employs a fine-grained region-query alignment module that enhances its attention to localized visual details, making it particularly adept at tasks involving small objects or dense scenes. Additionally, its multilingual comprehension and instruction-following capability broaden its applicability in global and real-world settings.

\paragraph{LLaVA.} LLaVA (Large Language and Vision Assistant) \citep{liu2023llava} integrates a vision encoder with the Vicuna language model, leveraging visual instruction tuning to align visual representations with natural language understanding. It has demonstrated strong performance on tasks such as Science QA, exhibiting capabilities comparable to those of multimodal GPT-4. We deploy \textbf{LLaVA-13B} locally.

LLaVA's core strength lies in its visual instruction tuning pipeline, where it is fine-tuned on multi-turn image-text instruction data. This strategy allows LLaVA to follow natural language prompts while reasoning over visual inputs effectively. Its architecture leverages pretrained weights from Vicuna for language and CLIP for vision, merging them through projection and alignment layers that maintain semantic coherence across modalities. This design empowers it to handle reasoning-heavy visual tasks with minimal fine-tuning.

\paragraph{Cleanlab.} Cleanlab \citep{northcutt2021confident} is an open-source Python library that implements the Confident Learning (CL) framework---a model-agnostic, data-centric approach for detecting and correcting noisy labels in machine learning datasets. Unlike traditional methods that primarily focus on adjusting model loss functions to handle noisy labels, Cleanlab addresses the root cause by estimating the joint distribution between noisy (observed) and true (latent) labels. This is achieved through three key principles: \textit{pruning}, to identify and remove noisy labels; \textit{counting}, to estimate noise rates using calibrated frequency statistics; and \textit{ranking}, to prioritize training examples based on their likelihood of being clean.

CL operates under the class-conditional noise assumption and uses model-predicted probabilities as input. By leveraging a data structure called the \emph{confident joint}, it robustly estimates the noise transition matrix and identifies mislabeled examples---even under class imbalance or imperfect probability calibration. Cleanlab supports a variety of learning paradigms, including multi-class and multi-label classification, and has demonstrated state-of-the-art performance in identifying real-world noisy labels in benchmark datasets such as CIFAR-10, ImageNet, and Amazon Reviews \citep{northcutt2021labelerrors}.

\paragraph{Docta.} 
Docta \citep{zhu2023unmasking} is an open-source framework for systematically auditing and improving the credibility of annotated language datasets, particularly in the context of safety alignment for large language models. It addresses the problem of mislabeled or inconsistent annotations that can undermine the reliability of downstream models, especially in tasks such as toxicity classification or safe response generation.

The core methodology of Docta is built upon estimating a label noise transition matrix and defining a data credibility metric that quantifies the alignment between noisy labels and their estimated true counterparts. Without requiring access to ground-truth labels, Docta leverages a $k$-NN label clusterability assumption and consensus-based soft labeling to identify noisy labels. A cosine similarity-based scoring function is then used to rank the likelihood of correctness for each instance, followed by threshold-based filtering to detect corrupted labels. 

\section{MTurk Annotation Format Example and Definition of Agreement Rate}

\subsection{MTurk annotation Format Example}
Taking CIFAR-10 as an example, the logic and format of the MTurk human annotation results\citep{northcutt2021labelerrors} are presented as follows:

\begin{center}
{\small
\begin{tcolorbox}[
  colback=blue!5!white, 
  colframe=gray!70!black, 
  title=MTurk examples on CIFAR-10,
  width=0.5\linewidth, 
  enlarge left by=0mm, 
  box align=center
]
\label{MTruk_examples_on_CIFAR-10}
\textbf{id}: 20 \\
\textbf{url}: https://labelerrors.com/static/cifar10/20.png\\
\textbf{given original label}: 7\\
\textbf{given original label name}: horse\\
\textbf{our guessed label}: 5\\
\textbf{our guessed label name}: dog\\
\vspace{0.5em}
\begin{tcolorbox}[colback=yellow!10, title=MTurk Results]
\textbf{given}: 3\\
\textbf{guessed}: 0\\
\textbf{neither}: 0\\
\textbf{both}: 2
\end{tcolorbox}
\end{tcolorbox}
}
\end{center}

In \citep{northcutt2021labelerrors}, the MTurk results are based on the Cleanlab framework. In their study, only those instances where Cleanlab identified a disagreement with the original label were submitted to MTurk for human evaluation. As a result, the MTurk outcomes are restricted to four possible categories: (1) \textbf{given}, referring to the original label of the image; (2) \textbf{guessed}, referring to the label suggested by Cleanlab; (3) \textbf{both}, indicating that both the original and Cleanlab-predicted labels are plausible; and (4) \textbf{neither}, indicating that neither is appropriate. While this review protocol facilitates targeted verification of label disagreements, it has a key limitation: it fails to adequately capture \textbf{missing label errors}, especially in scenarios where the number of plausible label candidates is large.

\subsection{Definition of Agreement Rate}
Based on the MTurk results, we define an \textbf{Agreement Rate} metric to evaluate the consistency between our method (including individual VLMs and the aggregated REVEAL framework) and human annotations.

To quantify the consistency between model predictions and human annotations, we define the \textbf{Agreement Rate} based on four distinct MTurk outcome types. Let $R_i$ denote the set of predicted labels for image $i$ by a given method (e.g., a VLM or REVEAL), and let $g_i$ and $s_i$ represent the \textbf{given} (i.e., original) and \textbf{guessed} (i.e., Cleanlab-predicted) labels for that image. Agreement is determined under the following conditions:

\begin{itemize}
    \item \textbf{Case 1 (MTurk = given):} Agreement is counted if $g_i \in R_i$.
    \item \textbf{Case 2 (MTurk = guessed):} Agreement is counted if $s_i \in R_i$.
    \item \textbf{Case 3 (MTurk = both):} Agreement is counted if both $g_i \in R_i$ and $s_i \in R_i$.
    \item \textbf{Case 4 (MTurk = neither):} Agreement is counted if $g_i \notin R_i$ and $s_i \notin R_i$.
\end{itemize}

We apply this agreement rule to five methods: four VLMs, namely \textbf{BLIP}, \textbf{LLaVA}, \textbf{Qwen}, and \textbf{Janus}, as well as our aggregated framework \textbf{REVEAL}. The overall agreement rate is defined as the proportion of images for which the prediction satisfies the corresponding MTurk-based agreement condition:

\[
\text{Agreement\_Rate} := \frac{1}{N} \sum_{i=1}^N \mathbb{I}\{\text{Image } i \text{ satisfies the agreement condition}\}
\]

where $N$ is the total number of evaluated images, and $\mathbb{I}\{\cdot\}$ is the indicator function that returns 1 if the condition holds and 0 otherwise.

\end{document}